# Intuitions about Ordered Beliefs Leading to Probabilistic Models

**Paul Snow**
Department of Computer Science
Plymouth State College
P.O. Box 6134
Concord, New Hampshire 03303-6134 USA

**Abstract**

The general use of subjective probabilities to model belief has been justified using many axiomatic schemes. For example, 'consistent betting behavior' arguments are well-known. To those not already convinced of the unique fitness and generality of probability models, such justifications are often unconvincing. The present paper explores another rationale for probability models. 'Qualitative probability,' which is known to provide stringent constraints on belief representation schemes, is derived from five simple assumptions about relationships among beliefs. While counterparts of familiar rationality concepts such as transitivity, dominance, and consistency are used, the betting context is avoided. The gap between qualitative probability and probability proper can be bridged by any of several additional assumptions. The discussion here relies on results common in the recent AI literature, introducing a sixth simple assumption. The narrative emphasizes models based on unique complete orderings, but the rationale extends easily to motivate set-valued representations of partial orderings as well.

## 1 INTRODUCTION

Many probabilists assert that subjective probabilities provide a generally useful foundation for the modeling of beliefs in all situations. That is, belief strengths are (or ought to be) realistically depicted as probability distributions, and that the strengths change in the face of evidence according to Bayes' formula.

Non-probabilists naturally take issue, some denying not only the generality of probability, but the generality of *any* single formalism (e.g. Smets 1991). Since there seems to be no controversy that probability models are sometimes useful, a natural question is where the frontiers of their usefulness might lie.

An old prescriptive argument that there are no frontiers, the *pignic* motivation (DeFinetti 1937, Savage 1972), appeals to consistency in betting behavior. This line of argument can be sidestepped fairly readily, as both Shafer and Zadeh have (in discussions of Lindley 1982), by noting that betting isn't the same as believing.

Another line of argument might be called *mathematical inevitability* (Cox 1946, Horvitz et al. 1986). Assuming that some scalar measure of belief strength exists, one enumerates mathematical properties that any such measure should have. The quality of this argument is quite different from the pignic. The point isn't so much that one ought to use probabilities, but rather that there is effectively no choice. Inevitability also opens the door to set-valued models (an ensemble of probability distributions not ruled out by the believer, advocated by such as Levi, 1980 and Kyburg 1987); the pignic position insists that there be only one probability distribution. Nevertheless, like any axiomatic system, inevitability can be countered by denying that one or more of the suggested mathematical properties is apodictic. Dubois and Prade (1988) have argued this position well.

A third approach observes that objective probabilities are known for some uncertain propositions. If the believer orders the propositions in one's corpus of beliefs, and some of those beliefs are represented by objective probabilities, then subjective probabilities can model all of the ordered beliefs. This notion receives axiomatic treatment in Villegas (1964) and DeGroot (1970). This approach can be defeated by asserting that the believer is simply unwilling, or computationally unable, to create such an ordering in all cases of interest.

This paper borrows ideas from all three approaches to arrive at a new and simple motivation for probabilism. Five basic assumptions about ordered beliefs are developed, leading to a motivation of 'qualitative probability' models. From there to ordinary probability run many paths often trod in an extensive literature. We shall explore an inevitability route, in the course of which a sixth assumption is introduced. A final note adapts the results developed for a single complete ordering of beliefs to models involving partial orderings.

## 2  ASSUMPTIONS FOR QUALITATIVE PROBABILITIES

Let us assume that the believer has an ordering $r()$ defined over a finite number of atomic sentences of interest and their Boolean combinations, including the always true sentence (T) and the always false sentence (F). We assume the following about $r()$:

A1. $r()$ is a complete transitive ordering.

A2. $r(T) > r(F)$.

A3. For all sentences **a** in the ordering,

$$r(T) >= r(a) >= r(F)$$

with equality obtaining if and only if **a** is asserted to be certainly true or certainly false.

A4. For any sentences **a** and **b**, and any **c** where

$$r(T) > r(c) > r(F)$$

if

$$r(a \mid c) >= r(b \mid c) \text{ and } r(a \mid \sim c) >= r(b \mid \sim c)$$

then

$$r(a) >= r(b)$$

and if either antecedent inequality is strict, then so is the consequent.

A5. For all sentences **a**, **b**, and **c**,

$$r(a \mid c) >= r(b \mid c)$$

if and only if

$$r(a \text{ and } c) >= r(b \text{ and } c).$$

In all these assumptions, the inequality symbols and the 'given stroke' are defined in the usual ways.

(Throughout, we shall consider the conditioning operation as invalid for 'given' sentences ordered equally with F). As stated, these assumptions lead to a discussion of belief models based upon a single belief ordering. In a later section, we shall modify these assumptions to support sets of orderings in the interest of motivating set-valued probability models.

The fourth assumption can be motivated as a kind of dominance principle, specialized here to the belief context rather than to the gambling context (but still within the spirit of Savage, 1972). The force of the argument is that since we believe that **a** is no less credible than **b** if **c** is true, and believe the same if **c** is false, and those are the only two possibilities, then our prior ordering should favor **a** over **b**.

The fifth assumption was taken from DeGroot (1970). In addition to the use we shall make of it here, A5 underlies DeGroot's motivation of Bayesian revision given that there is some probability distribution $p()$ which represents $r()$. Indeed, A5 is a straightforward statement of the Bayesian intuition about what it means for evidence to favor one proposition over another.

**Theorem.** For sentences **a**, **b**, and **c** where **a and c** and **b and c** are both false, then

$$r(a \text{ or } c) >= r(b \text{ or } c)$$

if and only if

$$r(a) >= r(b).$$

**Proof.** Note that if **c** is asserted to be false, then the equivalence is immediate.

(1) From the contrapositive of the weak inequality portion of A4 and from the easily derived $r(a \text{ or } c \mid c) = r(b \text{ or } c \mid c) = r(T)$, it follows that if $r(a \text{ or } c) > r(b \text{ or } c)$, then $r(a \text{ or } c \mid \sim c) > r(b \text{ or } c \mid \sim c)$, and thus, by application of the assumption A5, $r((a \text{ or } c) \text{ and } \sim c) > r((b \text{ or } c) \text{ and } \sim c)$, or simply $r(a) > r(b)$.

(2) Similarly, by the strict inequality portion of A4, if $r(a \text{ or } c) = r(b \text{ or } c)$, then $r(a \text{ or } c \mid \sim c)$ cannot be strictly greater nor strictly less than $r(b \text{ or } c \mid \sim c)$, and so by A5 again, we conclude $r(a) = r(b)$.

(1) and (2) are the 'only if' portion; the contrapositive of (1) is the 'if' portion. //

The usual definition of *qualitative probability* comprises A1, A2, A3, and the theorem. Thus, we conclude that ordering $r()$ is a qualitative probability. We now seek conditions under which there exists some probability distribution $p()$ which *agrees* with $r()$, that is, for all sentences **a**, **b** in the ordering, r (**a**) >= r (**b**) if and only if p (**a**) >= p (**b**).

## 3 FROM QUALITATIVE TO ORDINARY PROBABILITY

Fishburn and Roberts (1989) review some of the literature on the additional assumptions needed for a qualitative probability on a finite domain to have an agreeing probability distribution. If there are four or fewer mutually exclusive and collectively exhaustive propositions of interest, then an agreeing probability distribution exists without further assumptions (Kraft et al., 1959). For practical work, of course, something more general is needed.

For an AI audience, it may be interesting to look at the question in light of the results of Horvitz, et al. (1986), which results are well-known in the AI literature. These authors showed that if a scalar measure suffices to represent the degree of belief (and an index on the finite ordering satisfies that), then the scalar is an increasing transformation of some probability distribution provided that, in our notation:

H1. For all sentences **a** and **b**,

r (**a**) >= r (**b**) implies r (~**a**) =< r (~**b**).

H2. For all sentences **a**, **b**, **c**, r (**a** and **b** | **c**) is a strictly increasing function of r (**a** | **b** and **c**) and of r (**b** | **c**) when the other is held constant.

Assumption H1, sometimes called *complementarity*, has been criticized on intuitive grounds by Dubois and Prade (1988). It is, however, proven that every qualitative probability exhibits complementarity (Kraft, et al., 1959).

Assumption H2, despite its relative complexity, is often presented as intuitively reasonable in its own right. However, Dubois and Prade (1990) have noted that if H2 is amended to read *weakly increasing* instead of *strictly increasing*, then the necessity of probability measures doesn't follow from H1 and the amended H2. This observation has great force if weakly increasing relationships serve the intuition underlying H2 as well as strictly increasing relationships do.

There is no way to resolve a clash of intuitions, but it may be helpful to derive H2 from other, simpler assumptions. That way, at least, further discussion is narrowed. We proceed by introducing a new assumption to the five already made, a specialized "chain rule":

A6. For all sentences **x**, **y**, and **z** in the ordering such that **x** implies **y** and **y** implies **z**, r (**x** | **z**) is a strictly increasing function of r (**x** | **y**) and of r (**y** | **z**) when the other is held constant.

We then note the following corollary of assumption A5:

r (**b** | **a**) = r (**b** and **a** | **a** )

Apart from its logical status, the corollary makes sense: if we learned that **a** was true, then our opinion about **b** would presumably coincide with our opinion about the conjunction of **b** and the now-known-to-be-true **a**. From these we offer the following proof of H2:

> **Proof.** In A6, let **x** = **a and b and c**, **y** = **b and c**, and **z** = **c**; **x** implies **y** implies **z** as required. The substitution into A6 asserts that r (**a and b and c** | **c**) is increasing in r (**a and b and c** | **b and c**) and in r (**b and c** | **c**). By the corollary, these three quantities can be rewritten to read, respectively, r (**a and b** | **c**), r (**a** | **b and c**), and r (**b** | **c**), which yields the text of H2.//

Thus, the present assumptions A1 through A6 imply Horvitz et al.'s H1 and H2, leading to the conclusion that there is at least one ordinary probability distribution which agrees with the ordering r ( ).

## 4 A NOTE ON SET-VALUED MODELS

While many find the notion of ordered belief intuitively plausible, many also find the notion of a *complete* ordering implausible. On computational grounds, there are a lot of beliefs to order. Further, there is no room for modesty in a complete ordering: perhaps the believer simply doesn't know enough to rank confidently the sentence 'it will rain tomorrow in Boston' against 'traffic will be heavy in Boston tomorrow night.' Indeed, a frequently-heard rejoinder to the pignic motivation of probability (and others which insist that only a single probability distribution will do) is that complete orderings are often psychologically unrealistic and pragmatically burdensome.

Only simple modifications of the earlier arguments are needed to support set-valued probability models of partially ordered beliefs. A convenient starting point for such a model is the set of all belief orderings not ruled out by either the beliefs actually held or by the assumptions A1 through A5.

Each member of the set would be a qualitative probability. If A6 (or any of the other 'bridging' assumptions studied in the literature) holds, then each member would be an increasing transform of one or more probability distributions. The final model would be the set of all probability distributions which agree with at least one ordering in the set of qualitative probabilities. Beliefs would be modeled by this arrangement in the sense of unanimous agreement, that is

$$r(a) >= r(b) \text{ in all orderings}$$

if and only if

$$p(a) >= p(b) \text{ in all distributions}$$

Such a model satisfies a simple generalization of the minimum plausible requirements for belief models identified by Prade (1985), namely

$$p(F) = 0$$
$$p(T) = 1$$

if $a$ implies $b$, then $p(a) =< p(b)$

since all probability distributions, and hence each distribution in the model set, display these properties. Of course, the distributions would not always agree on particular values for $p(a)$ and $p(b)$, although they would agree on the inequality; hence the need to somewhat generalize Prade's criteria.

## 5 CONCLUSIONS

There is no possibility that the enumeration of any set of axioms will confound the non-believers, nor is it necessary to the success of the probabilist program that other scholars quietly leave the field. On the contrary, the criticisms of Zadeh, Dempster and Shafer, and many others of like mind spur the development and evolution of probabilist ideas. Probabilists who favor set-valued models, for instance, owe a debt to these non-probabilists for their critique of the pignic approach to belief modelling.

One use of an axiomatic exercise is to get beyond theories of 'brainwashing' as the source of probabilist scholarship, and to move the debate on to more profitable subjects. For example, axioms can be helpful when some uncertainty calculus yields a distinctly different answer from what any probability model would yield. Does a probabilist have a license to object that the other calculus is 'wrong'? Axioms have a role in clarifying what is at stake in a difference between approaches, even if axioms cannot settle any clash of intuitions that the difference reveals.

Even so, it must be admitted that the lively rivalry between the probabilists and some others may be much ado about the difference between two wrong answers. Anand (1987), for one, questions the generality of the notion of ordered belief. Perhaps what is really going on sometimes, he argues, is more like an athletic league. 'Teams' (sentences) compete for top honors, and any ordering is transitory with little relevance to how future 'contests' (i.e., what will happen when new evidence is seen) will turn out.

Further, there may be questions about the experience of uncertainty that go beyond the ordering of sentences, even if it is uncontroversial that such an ordering exists. For example, Shackle (1949) considers the cognitive phenomenon of *surprise*. While it may be the case that probabilistic models of such a phenomenon can be built, and that probability principles may constrain what models are acceptable, nevertheless, the phenomenon goes beyond the issues usually addressed by probability. Thus, it is unsurprising that Shackle proposes a model which lacks any explicit probability content.

More recently, Hsia (1991) has modeled surprise using the Dempster-Shafer formalism. Hsia's work is especially interesting because he asks whether D-S might be capturing an aspect of the experience of uncertainty which is distinct from the aspect on which Bayesians usually focus.

Let us finish by returning to the question of where the frontiers of probabilism's usefulness might lie, first raised in the introduction. The gist of the axioms presented here is that probability models are appropriate whenever (1) the believer subscribes to certain constraints about how sentences may be ordered in 'belief-worthiness', (2) the believer subscribes to certain constraints about how such an ordering would change if evidence were observed, and (3) the objective of the model is to represent the orderings that are subject to these constraints and to constraints imposed by the actual beliefs themselves.

No claim is made that the constraints of (1) and (2) are shared by all believers with respect to all beliefs, nor that they ought to be. No claim is made that all aspects of the experience of uncertainty are necessarily manifestations of such a constrained belief ordering. What is claimed is that many believers do acknowledge such constraints, and that many aspects of the uncertainty experience are manifestations of an ordering of beliefs.

What the axioms allow, then, is a formal demonstration that for such a believer and for a question about such an aspect of uncertainty, a probability model is required to do justice to both the constraints and the beliefs. (That does not rule out, of course, the possibility that some other method might furnish satisfactory service as an approximation tool.) A further goal is to state the constraints simply, so that discussion about what a believer subscribes to, and about what the problem in question is about, can proceed easily.

Axioms like those presented here serve to stake out a claim about the scope of a method. Whether that amounts to all occasions of uncertain reasoning or not, simple and explicit articulation of the boundaries of a claim is desirable in itself. As the poet Robert Frost remarked, good fences make good neighbors.